\documentclass[acmtog]{acmart}

\usepackage{booktabs} 

\usepackage[utf8]{inputenc} 
\usepackage[T1]{fontenc}    
\usepackage{hyperref}       
\usepackage{url}            
\usepackage{booktabs}       
\usepackage{amsfonts}       
\usepackage{nicefrac}       
\usepackage{microtype}      
\usepackage{xcolor}         
\definecolor{myPurple}{rgb}{0.4, .0, .8}
\definecolor{myGreen}{rgb}{0.2, .8, .3}
\definecolor{myRed}{rgb}{0.8, .2, .2}
\definecolor{myOrange}{rgb}{0.8, 0.45, 0.0}
\definecolor{myBlue}{rgb}{.0, .0, 1.0}

\usepackage{xcolor}
\usepackage{colortbl}
\definecolor{color3}{RGB}{255, 255, 200}
\definecolor{color2}{RGB}{255, 220, 200}
\definecolor{color1}{RGB}{255, 181, 163}
\newcommand{\cc}[1]{\cellcolor{color#1}}
\usepackage{enumitem} 

\usepackage[ruled]{algorithm2e} 

\acmJournal{TOG}

\begin{document}
\title{Anti-Aliased Neural Implicit Surfaces with Encoding Level of Detail}

\author{Yiyu Zhuang}
\authornotemark[2]
\affiliation{%
 \institution{Najing University}
 \city{Nanjing}
 \country{China}}
\email{yiyu.zhuang@smail.nju.edu.cn}
\author{Qi Zhang}
\authornotemark[2]
\affiliation{%
 \institution{Tencent AI Lab}
 \city{Shenzhen}
 \country{China}
}
\email{nwpuqzhang@gmail.com}
\author{Ying Feng}
\affiliation{%
 \institution{Tencent AI Lab}
 \city{Shenzhen}
 \country{China}
}
\email{yfeng.von@gmail.com}
\author{Hao Zhu}
\affiliation{%
 \institution{Najing University}
 \city{Nanjing}
 \country{China}}
\email{zhuhaoese@nju.edu.cn}
\author{Yao Yao}
\affiliation{%
 \institution{Najing University}
 \city{Nanjing}
 \country{China}}
\email{yyaoag@cse.ust.hk}
\author{Xiaoyu Li}
\affiliation{%
 \institution{Tencent AI Lab}
 \city{Shenzhen}
 \country{China}
}
\email{xliea@connect.ust.hk}
\author{Yan-Pei Cao}
\affiliation{%
 \institution{Tencent AI Lab}
 \city{Shenzhen}
 \country{China}
}
\email{caoyanpei@gmail.com}
\author{Ying Shan}
\affiliation{%
 \institution{Tencent AI Lab}
 \city{Shenzhen}
 \country{China}
}
\email{yingsshan@tencent.com}
\author{Xun Cao}
\affiliation{%
 \institution{Najing University}
 \city{Nanjing}
 \country{China}}
\email{caoxun@nju.edu.cn}
\renewcommand\shortauthors{Zhuang et al.}

\begin{abstract}
    \footnotetext[2]{Both authors contributed equally to this work. Zhuang did this work during the internship at Tencent AI Lab mentored by Zhang.}
    We present LoD-NeuS, an efficient neural representation for high-frequency geometry detail recovery and anti-aliased novel view rendering. Drawing inspiration from voxel-based representations with the level of detail (LoD), we introduce a multi-scale tri-plane-based scene representation that is capable of capturing the LoD of the signed distance function (SDF) and the space radiance. 
    Our representation aggregates space features from a multi-convolved featurization within a conical frustum along a ray and optimizes the LoD feature volume through differentiable rendering. 
    Additionally, we propose an error-guided sampling strategy to guide the growth of the SDF during the optimization. 
    Both qualitative and quantitative evaluations demonstrate that our method achieves superior surface reconstruction and photorealistic view synthesis compared to state-of-the-art approaches.

\end{abstract}

%
%


%
%


\begin{teaserfigure}
    \centering
    \includegraphics[width=\linewidth]{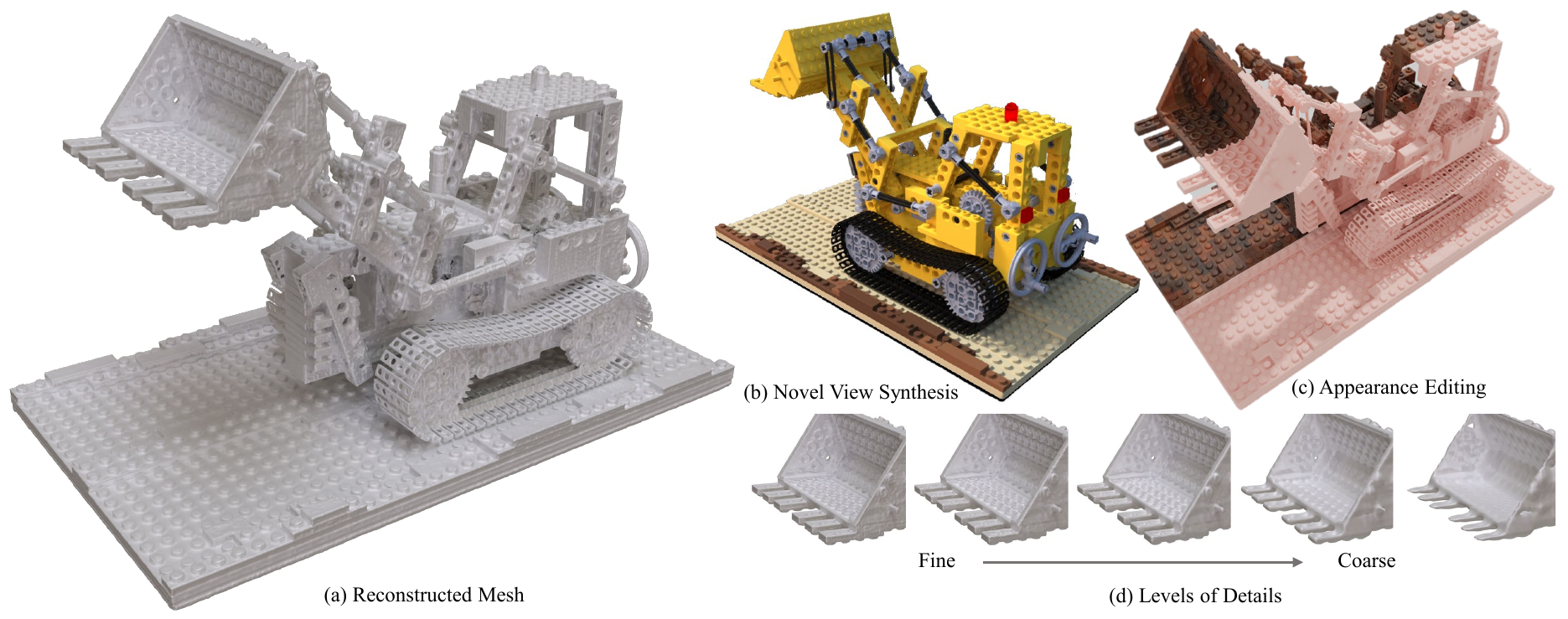}
    \caption{Our method, called \textit{LoD-NeuS}, adaptively encodes Level of Detail (LoD) features derived from the multi-scale and multi-convolved tri-plane representation. By optimizing a neural Signal Distance Field (SDF), our method is capable of reconstructing high-fidelity geometry (a). LoD-NeuS effectively captures varying levels of detail (d), resulting in anti-aliasing reconstruction, and thus, enabling photorealistic view synthesis (b) and appearance editing (c). }
    \label{fig:my_label}
\end{teaserfigure}

\maketitle

\section{Introduction}

Recent advances in implicit representation and neural rendering (i.e., NeRF~\cite{mildenhall2020nerf} approaches) have provided a new alternative for geometric modeling and novel view rendering. {However, applying the vanilla NeRF with the soft density representation to accurately reconstruct the geometry with fine-grained surface details remains challenging. In contrast, the neural implicit surface (NeuS)~\cite{wang2021neus} was proposed to apply the signed distance function (SDF) rather than the soft density to model the object surface within the NeRF framework explicitly.}
The object surface is represented as the zero-level set of the SDF modeled by the multi-layer perceptron (MLP).
NeuS and its variants have shown that SDF can flexibly represent the scene geometry with arbitrary topologies, and produce significantly better results in neural surface reconstruction than the vanilla NeRF approach. 

One of the major challenges of neural surface reconstruction is the reconstruction of high-frequency surface details. While frequency position encoding has been employed in NeuS, it still struggles to capture fine-grained geometry details accurately, resulting in low-fidelity and over-smooth geometric approximations for intricate models. HF-NeuS \cite{wang2022hf} attempts to mitigate the issue by introducing a displacement network tailored specifically for learning high-frequency geometry details. 
However, the problem persists due to the inherent limitations of frequency position encoding, which lacks locality and fails to adaptively capture different level of detail (LoD) in surface geometry. 
Consequently, the undersampling and inadequate representation of high-frequency information inevitably results in aliasing artifacts during novel view rendering.

On the other hand, explicit voxel-based representations have long employed multi-scale prefiltering techniques, such as mipmaps and octrees, to enable fine-grained surface recovery and anti-aliasing in object rendering. Recent advancements in NeuS-based approaches have also explored the potential of hybrid implicit-explicit representations. These methods replace multi-layer perceptrons (MLPs) with discretized volumetric representations, such as voxel grids \cite{yu2022monosdf} and tri-planes \cite{wang2023petneus}, resulting in better geometric approximations. However, these methods possess inherent limitations (Sec. 2), which pose challenges when combining the anti-aliasing advantages of explicit methods with hybrid representations in surface reconstruction with continuous LoD.

In this paper, we present a neural implicit surface representation with the encoding level of detail (\textit{LoD-NeuS}) for high-quality geometry reconstruction from multi-view images. The implicit surface is represented by a multi-scale tri-plane-based feature volume, which is optimized through differentiable cone sampling and volume rendering.
\begin{enumerate}
    \item We present a tri-plane position encoding, optimizing multi-scale features, to effectively capture different levels of detail;
    \item We design a multi-convolved featurization within a conical frustum, to approximate cone sampling along a ray,    {which enables the anti-aliasing recovery with finer 3D geometric details;}
    \item We develop a refinement strategy, involving error-guided sampling, to facilitate SDF growth for thin surfaces.  
\end{enumerate}
In experiments, our method outperforms state-of-the-art NeuS-based approaches at high-quality surface reconstruction and view synthesis, particularly for objects and scenes with high-frequency details and thin surfaces.

\section{Related Work}
\noindent\textbf{Multi-view 3D Reconstruction. }
Reconstructing the surfaces of the scene from multi-view images is a fundamental problem that has been extensively studied throughout the development of computer vision and graphics \cite{han2019image}. Multi-view 3D reconstruction has three categories: point-based reconstruction \cite{barnes2009patchmatch, campbell2008using, furukawa2009accurate, tola2012efficient, schonberger2016pixelwise}, surface reconstruction \cite{hoppe1992surface, kazhdan2006poisson, izadi2011kinectfusion, dai2017bundlefusion},  and volumetric reconstruction \cite{de1999poxels, seitz1999photorealistic, kutulakos2000theory, broadhurst2001probabilistic}. Traditionally, point-based or surface-based methods first  estimate the geometry information (\textit{e.g.}, depth and normal maps) of each pixel by matching the correspondences of multi-view images \cite{schonberger2016structure}, and then fuse the geometry information \cite{merrell2007real, zach2007globally} followed by mesh surface reconstruction processes such as Delaunay triangulation \cite{labatut2007efficient} and ball-pivoting \cite{bernardini1999ball}. 
The performance of surface reconstruction largely depends on the accuracy of correspondence matching. Recovering surfaces with minimal textures can be challenging, leading to significant artifacts and partially missing reconstructed content. To circumvent issues with insufficient geometry correspondence, volumetric methods \cite{niessner2013real, sitzmann2019deepvoxels} estimate occupancy and color within a voxel grid from multi-view images and evaluate color consistency at each grid. Nevertheless, these volumetric approaches involve explicitly breaking down a scene into a vast number of samples, which necessitates substantial storage capacity, thereby limiting grid resolution and impacting the overall reconstruction quality.

\noindent\textbf{Neural Implicit Surface.} 
Recent advances in implicit neural representations have showcased the potential to reconstruct highly detailed surfaces and render photorealistic views \cite{tewari2022advances}. Neural Radiance Fields (NeRF) \cite{mildenhall2020nerf}, a notable breakthrough in this domain, learns the radiance fields (density and view-dependent color) of a scene and renders novel views based on volumetric ray tracing. NeRF and its variations have been applied to a range of tasks, including novel view synthesis \cite{barron2022mip, chen2022hallucinated, zhu2023pyramid}, generalizable models \cite{zhuang2022mofanerf, huang2023lirf, wu2023describe3d}, imaging processing \cite{huang2022hdr, ma2022deblur, huang2023inverting}, and inverse rendering \cite{srinivasan2021nerv, verbin2022ref, zhuang2023neai}.
However, compared to the signed distance function (SDF) \cite{chabra2020deep, genova2020local} or occupancy field \cite{mescheder2019occupancy, oechsle2021unisurf}, recovering smooth and accurate surfaces using the density function is challenging, often produce noisy low-fidelity geometry approximation since it lacks sufficient constraints on its level sets.
Specifically, VolSDF \cite{yariv2021volume} incorporates an SDF into the density function, ensuring that it satisfies a derived error bound on the transparency function. 
NeuS \cite{wang2021neus} proposes an unbiased formulation with a logistic sigmoid function and introduces a learnable parameter to control the slope of the function during the rendering and sampling processes. Building upon NeuS, NeuralWarp \cite{darmon2022improving} and Geo-NeuS \cite{fu2022geo} leverage prior geometry information from MVS methods but may struggle in the regions with less texture. HF-NeuS \cite{wang2022hf} integrates additional displacement networks to fit the high-frequency details. However, the frequency position encoding used in these methods struggles to adaptively capture varying levels of detail (LoD) across different regions. Additionally, using ray sampling instead of cone sampling leads to undersampled or inaccurately represented high-frequency information, which finally results in aliasing artifacts.

\noindent\textbf{Anti-aliased Representation.} 
Surfaces employing traditional explicit representations (\textit{e.g.}, polygon mesh, voxel grids), can be efficiently reconstructed without encountering aliasing artifacts, thanks to the application of multi-scale prefilter techniques. 
These techniques, such as mipmaps and octrees, offer a robust solution for handling different levels of detail in surfaces while maintaining efficiency.
Continuous implicit surface representations achieve higher performance but can only be anti-aliased through supersampling, which further slows down their already time-consuming reconstruction. The hybrid explicit-implicit representation emerges as a result.
In particular, \citeN{takikawa2021neural} proposes a multi-scale representation based on sparse voxel octrees for implicit surface with a learned geometry prior. 
MonoSDF \cite{yu2022monosdf} employs a multi-scale voxel-based representation with monocular geometric cues for SDF reconstruction, which introduces the hash encoding \cite{muller2022instant} to optimize the grid feature.
Although hash encoding can enhance both memory efficiency and performance, it potentially causes hash collisions and representations that are insufficiently explicit.
PET-NeuS \cite{wang2023petneus} adopts a self-attention convolution to generate the tri-plane-based representation \cite{chen2022tensorf, chan2022efficient} for enhancing quality, but following positional encoding on both tri-plane features and position increase model parameters and computational complexity. Besides, the performance of PET-NeuS heavily depends on the effectiveness of self-attention.
Inspired by these ideas, we introduce an efficient neural representation to aggregate LoD features, for the first time, that enables continuous LoD with cone sampling while achieving state-of-the-art geometry reconstruction quality.

\section{Preliminaries}
This section overviews the base priors of NeRF \cite{mildenhall2020nerf} for volume rendering, as well as geometric improvements extended by SDF and NeuS \cite{wang2021neus} for surface reconstruction and view synthesis. 

\textbf{NeRF} represents the scene with a continuous volumetric radiance field, which utilizes MLPs to map the position  $\mathbf{x}$ and view direction $\mathbf{r}$ to a density $\sigma$ and color $\mathbf{c}$. To render a pixel's color, NeRF casts a single ray $\mathbf{r}(t)=\mathbf{o}+t\mathbf{d}$ through the pixel and samples a set of points with different $\{t_i\}$ along the ray. The evaluated $\{(\sigma_i, \mathbf{c}_i)\}$ at the sampled points are accumulated into the color ${C}(\mathbf{r})$ of the pixel via volume rendering \cite{max1995optical}:
\begin{equation}\small
     {C(r)}\!=\!\sum_{i} T_i \alpha_i \mathbf{c}_i, 
     where\  T_i = \exp  \left({-\sum_{k=0}^{i-1}\sigma_{k}\delta_{k}}\right),
    \label{eq:vol_ren} 
\end{equation}
and $\alpha_i=1-\exp(-\sigma_i\delta_{i})$ indicates the opacity of the sampled point. Accumulated transmittance $T_i$ quantifies the probability of the ray traveling from $t_0$ to $t_i$ without encountering other particles, and $\delta_{i} = t_i-t_{i-1}$ denotes the distance between adjacent samples. 

\textbf{NeuS} extends the basic NeRF formulation by integrating an SDF into volume rendering. 
It represents the scene's geometry with a learnable function $f$, which returns the signed distance $f(\mathbf{x})$ from each point to the surface. The underlying surface can be derived from the zero-level set,
\begin{equation}\small
S=\{\mathbf{x}\in\mathbb{R}^3|f(x)=0\}.
\end{equation}
Subsequently, NeuS defines a function to map the signed distance to density $\sigma$, which attains a locally maximal value at surface intersection points. Specifically, accumulated transmittance $T(t)$ along the ray $\mathbf{r}(t)=\mathbf{o}+t\mathbf{d}$ is formulated as a sigmoid function: $T(t)=\Phi(f(t))=(1+e^{sf(t)})^{-1}$, where $s$ and $f(t)$ refers to a learnable parameter and the SDF value of point at $\mathbf{r}(t)$, respectively. Discrete opacity values $\alpha_i$ can then be derived as:
\begin{equation}\small
\alpha_i=max(\frac{\Phi_s(f(t_i)) - \Phi_s(f(t_{i+1}))}{\Phi_s(f(t_i))}, 0).
\label{eq:neus_opacity}
\end{equation}
NeuS employs volume rendering to recover the underlying SDF based on Eqs. \eqref{eq:vol_ren} and \eqref{eq:neus_opacity}. The SDF is optimized by minimizing the photometric loss between the renderings and ground-truth images.

\section{Method}

Drawing inspiration from anti-aliasing techniques for explicit voxel-based surface reconstruction, we aim to develop a hybrid representation that combines the advantages of both explicit and implicit representations, to achieve anti-aliasing and the recovery of delicate geometric details. In particular, we firstly present a novel position encoding based on multi-scale tri-planes to enable continuous levels of details (Sec. \ref{sec:implicit_Surface_Representation}). To alleviate aliasing, we consider the size of cast cone rays (similar to \cite{barron2022mip}) and specifically design multi-convolved features to approximate the cone sampling (Sec. \ref{sec:anti-rendering}). Meanwhile, we observe that thin surface reconstruction using SDF is challenging, thus propose a refined solution involving an error-guided sampling strategy to facilitate SDF growth (Sec. \ref{sec:growth}).

\subsection{Multi-scale Tri-plane Encoding}
\label{sec:implicit_Surface_Representation}
 
Recent progress \cite{yu2022monosdf, muller2022instant} in the field of neural rendering has shown that incorporating learnable features extracted from multi-scale grids significantly enhances reconstruction quality and accelerates volume rendering. In contrast to voxel-based representations with heavy memory requirements and hash encoding with collision issues, tri-plane-based representations \cite{chan2022efficient, chen2022tensorf} provide increased flexibility in handling complex geometry and effective spatial regularization. Inspired by these insights, we incorporate the multi-scale tri-plane representation into a NeuS-based framework for intricate surface reconstruction and high-quality rendering.

To address the challenges associated with reconstructing high-frequency details and achieve a more reasonable implicit surface representation, we propose a learnable encoding based on multi-scale tri-planes. A tri-plane representation $\mathbf{P}$ is a novel 3D data structure, which consists of three learnable feature planes $\{P_{xy},{P}_{xz},{P}_{yz}\}$. These planes are orthogonal to each other and form a 3D cube centered at the origin $(0,0,0)$. For each 3D point $\mathbf{x}\in\mathbb{R}^3$, we project it onto each of the three planes, gathering features ${F}_{xy},{F}_{xz},{F}_{yz}$ using bilinear interpolation. 
The element-wise concatenation of these features yields the feature $\mathbf{F}$ with dimensionality $N$. 

Unlike previous methods \cite{chan2022efficient, chen2022tensorf}, we construct a set of tri-planes with different resolution $\{R_l\}^L_{l=1}$, where $L$ indicates the number of levels. Each level is independent and stores feature at the vertices of tri-plane. Our position encoding function, $\gamma(\mathbf{x})$, concatenates the input $\mathbf{x}$ with the feature $\mathbf{F}_l$ from every level $l$, forming a multi-scale feature vector $\vec{\mathbf{F}}=(\mathbf{x},\mathbf{F}_1,...,\mathbf{F}_L)$, whose length is $3 + L \times N$. 
By replacing the traditional frequency position encoding with the multi-scale tri-plane feature vector, our method benefits from explicit representation while guarantees different levels of detail.

\subsection{Anti-aliasing Rendering of Implicit Surfaces}
\label{sec:anti-rendering} 
Once we have acquired multi-scale tri-plane features, our goal is to estimate the SDF of samples along a ray for volume rendering. NeuS renders a pixel's color by casting a single ray through the pixel, without considering its size and shape. This approximation potentially leads to undersampling or ambiguous representation of high-frequency information, and results in aliasing artifacts. To alleviate this, we reformulate volume rendering by defining a ray as a cone, taking into account the pixel size. This enables the continuous LoD and recovers a high-quality SDF from undersampled images, leading to more accurately capture and reconstruction of the fine details of the scene.

A straightforward solution is to discretize a cone into a batch of rays, similar to super-sampling techniques \cite{cook1986stochastic}. However, this approach increases the number of sampled rays and points for volume rendering, leading to prohibitively high computational costs and slow inference time. 

\begin{figure} [t]
    \centering
    \includegraphics[width=1.\linewidth]{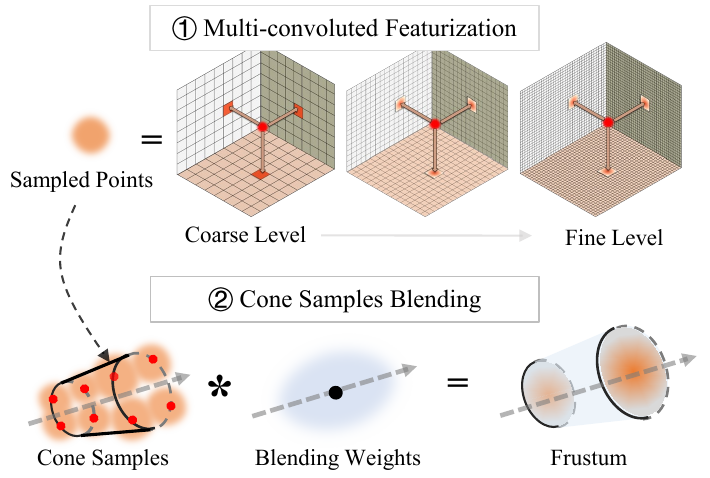}
    \caption{Aggregation of LoD feature, including multi-convolved featurization and cone discrete sampling. We obtain the feature of any sample within the conical frustum by blending the features of vertices. Additionally, considering the size of the sampled points, we introduce multi-convolved features by Gaussian Kernel to efficiently represent ray sampling within a cone. Combining both of them, we aggregate the LoD feature of any sample in a continuous manner.}
    \label{fig:anti-alising-statement}
\end{figure}

Here we provide a more efficient solution, including a sampling strategy and featurization procedure, in which we cast a cone and integrate features within conical frustums, as shown in Fig. \ref{fig:anti-alising-statement}.
\paragraph{\textbf{Cone Discrete Sampling.}}
Assuming a casting cone ray through a camera pixel is divided into a series of conical frustums, we need to integrate the color and geometry information within each conical frustum. Drawing inspiration from Mip-NeRF \cite{barron2022mip}, we attempt to integrate all features rather than just the network output. Note that, our target differs from Mip-NeRF, which focuses on rendering scenes at different resolutions rather than recovering scene details. 
{Utilizing our tri-plane-based representation, we cast four additional rays through the pixel corners, thus takes the pixel size and shape into account.}
Each conical frustum along the cone is then represented by eight vertices. 
Given any 3D sampled position $\mathbf{x}$ within a conical frustum, we blend the tri-plane features of each vertex $\mathbf{x}_v$ using decreasing weights, 
\begin{equation}\small
    W(\mathbf{x},\mathbf{x}_v) = \exp(-k |\mathbf{x}_v-\mathbf{x}|),
    \label{eq:blend_weight}
\end{equation}
which decreases with the distance between the vertex $\mathbf{x}_v$ and the sampled point $\mathbf{x}$. {$k$ is a learnable parameter that we initially set to $80$ and update along with other parameters during traing.} It is important to note that the decreasing function should be aware of the size of the conical frustum. The smaller the conical frustum is, the more rapidly the function should decrease. 

\paragraph{\textbf{Mulit-convolved Featurization.}} 
Though the multi-scale features of neighbor vertices along neighbor rays are applied for cone sampling, this approximation may be insufficent due to the sparse samples within the conical frustum. A straightforward way is to introduce more discretized samples, but this increases the computational cost and memory burden. Fortunately, the proposed explicit tri-plane-based representation makes it easy to integrate the features. Our approach utilizes the 2D Gaussian of each tri-plane to represent the region where the conical frustum should be integrated. In conjunction with our cone discrete sampling, we propose a multiple Gaussian convolved featurization to represent the features of neighbor vertices that approximate the sampled point and its corresponding conical frustum. Specifically, given a vertex $\mathbf{x}_v$, we project it onto the tri-planes and query the corresponding multi-scale feature vector $\vec{\mathbf{F}}_v$. Considering the grid resolution of the tri-plane, we apply multiple Gaussian convolutions with different kernel sizes. It represents the feature aggregation of samples within the conical frustum in a continuous manner. The multi-scale multi-convolved feature $\mathbf{G}_v$ for each vertex of the conical frustum is defined as: 
\begin{equation}\small
    \mathbf{G}_v(\mathbf{x}_v) = \mathcal{G}(\vec{\mathbf{F}}_v, \left\{\tau_v\right\}_{l=1}^{L})
    =\sqcup_{l=1}^{L}\mathcal{G}\left(\mathbf{F}_{l}, \tau_l\right),
    \label{eq:2d_gaussian}
\end{equation}
{
where $\mathcal{G}(\mathbf{F}, \tau)$ refers to our Gaussian convolution defined by covariance $\tau$. 
Through the 2D convolution, which aggregates the features of the compressed planes, we gather the features within a 3D sphere centered at $x_v$. 
We choose different kernel sizes $\{\tau_l\}$ for each level $l$ to covers various frequency details. The convolved features are combined using the concatenation operation, denoted as $\sqcup$.}
{Consequently, similar to position encoding, our featurization is able to cover different frequency details of the scene.
}

According to Eqs. \eqref{eq:blend_weight} and \eqref{eq:2d_gaussian}, for a sample at position $\mathbf{x}$ within the corresponding conical frustum, its LoD feature with continuous levels of detail is defined as,
\begin{equation}\small
    \mathbf{Z}(\mathbf{x})=\sum_{v=1}^{V}{W(\mathbf{x}, \mathbf{x}_v)\mathbf{G}_v(\mathbf{x}_v)},
\end{equation}
where $V=8$ is the number of vertices of a conical frustum. 
{
Compared to utilizing the 3D shape-adaptive Gaussian kernel to represent an ideal approximation of the cone discrete sampling, our formulation represents the solution spaces with a single leanable parameter $k$ rather than dealing with the complexity of shape-adaptive Gaussian kernels. }

\subsection{Training and Loss}
\label{sec:train}
After obtaining LoD feature $\mathbf{Z}$ of the samples along a ray, the colors and signed distance $f$ can be predicted. We do this via a shallow $8$-layer MLP: $(f, \Theta)=\mathrm{MLP}(\mathbf{Z})$. In addition to $f$, it produces a feature vector $\Theta\in\mathbb{R}^{256}$, which is then passed to the color module. According to Eq. \eqref{eq:neus_opacity}, we obtain the opacity $\alpha$ of the sampled point. The color module is represented as a $3$-layer MLP, which predicts the color $\mathbf{c}$ from $\Theta$ and view direction $\mathbf{d}$ as $\mathbf{c}=\mathrm{MLP}_c(\Theta, \mathbf{d})$. We finally follow Eq. \eqref{eq:vol_ren} to render pixel color $C_p$.

The learnable parameters and networks are optimized by employing a loss function and the process of backward propagation. To be more specific, a batch of $n$ pixels are randomly sampled, including their color $\{C_p\}_{p=1}^n$ and optional masks $\{M_p\}_{p=1}^n$. We further sample $m$ points along each ray, yielding the predicted color $\{\hat{C}_p\}$. Then the L1 loss is calculated to measure the reconstruction distance, which is defined as:
\begin{equation}\small
    L_{rgb}=\frac{1}{n}\sum_p \left \|  
            \hat{C_p} - C_p
    \right \|_1.
\end{equation}
We also add an Eikonal term \cite{gropp2020implicit} on all sampled points $\{x_i\}_{i=1}^{nm}$ to regularize the SDF by:
\begin{equation}\small
    L_{eikonal}=\frac{1}{nm}\sum_i (
            \left \|  
                    \nabla f(x_i)
            \right \|_2 
            - 1)^2.
\end{equation}
The mask loss $L_{mask}$ is optional and defined as:
\begin{equation}\small
    L_{mask}=\frac{1}{n}\sum_p  \mathrm{BCE}(M_p,\hat{O}_p),
\end{equation} 
where $\hat{O}_k\!=\!\sum_{j}^m T_j ( 1-\exp(-\sigma_j\delta_{j}) )$ is the opacity accumulated along the ray, and $\mathrm{BCE}$ is the binary cross entropy loss \cite{wang2021neus}.

\begin{figure*}[tb]
    \centering
    \includegraphics[width=1.0\linewidth]{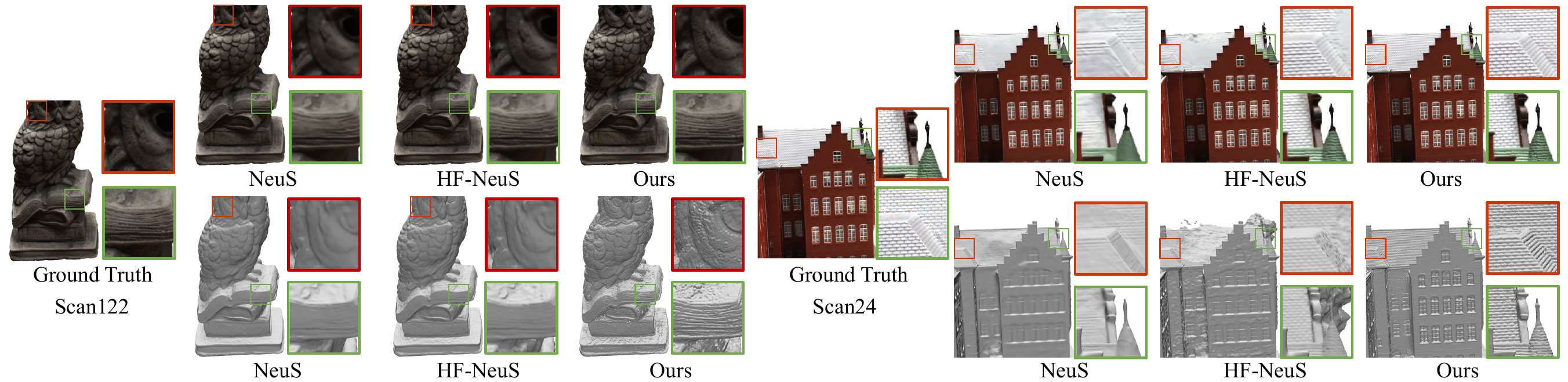}
    \vspace{-0.25in}
    \caption{Qualitative comparison with zoom-in details of our method against baselines on DTU dataset. Our method produces the most visually pleasing novel views and reconstructed geometry, especially on the intricate details and embossed patterns on the sculptures (left), along with the uneven surface created by the roof tiles (right), as shown in zoom-in details. 
    }
    \label{fig:comp_dtu}
\end{figure*}

\begin{algorithm}[tb] 
\label{alg:sdf_grow}
\caption{SDF Growth Refinement}\label{alg1}
\KwIn{SDF $f$, Rendered image $I_r$, Input image $I$, Step $n$}
\KwOut{Refined SDF $f_r$}
$e \gets \mbox{\it CalculateErrorMap}_{L1}(I_r, I)$\;
$M_e \gets \mbox{\it EstimateGrowthPoint}(I)$\;
$M_s \gets \mbox{\it ExpandRegion }(e)$\;
\For{$\mathrm{step} \in \{1, \dots, n\}$}{
    $M_e \gets \mbox{\it ExpandRegion}(M_e) \cap M_s$\;
    $M_s \gets M_s \setminus M_e$\; 
    $mask\_sequence.append(M_e)$;
}
\For{$\mathrm{step}, mask \in enumerate(mask\_sequence)$}{
    $sampled\_rays \gets \mbox{\it SampleRaysInsideMask}(mask)$\;
    $f \gets \mbox{\it OptimizeSDF}(f, sampled\_rays)$\;
}
$f_r \gets f$\;
\end{algorithm}

\subsection{SDF Growth Refinement}
\label{sec:growth}
We have observed that the SDF faces challenges when reconstructing thin objects, mainly for two reasons. First, representating a thin object necessitates a rapid flip in the SDF, which is difficult for the neural network \cite{yu2022monosdf}. Second, the image area corresponding to the thin object may have fewer samples compared to other areas, making it harder to learn. 
Based on this observation, a straightforward solution might be to increase the sampling frequency around this area. However, it's important to note that the optimization process of SDF is fundamentally different from NeRF. In NeRF, the optimization runs in a spatially-independent manner, which means changes at one point does not affect another. 

In contrast, the SDF represents the signed distance to the whole surface, which implies that when the implicit surface changes, the SDF values in a region are affected.
Due to this interconnected nature, the optimization process of the SDF appears to deform the initial geometry to better fit the target surface. This property helps maintain connectivity and prevents floating artifacts. However, this also complicates the reconstruction of thin objects, as only the sampled rays located around the region are proven helpful for this reconstruction.

To utilize the property, we devise a strategy to refine the optimized SDF for better thin object reconstruction, as shown in Alg. \ref{alg1} and Fig. \ref{fig:refinement}. Our motivation is to guide the SDF growth from the spatial point where the missing thin segment meets the surface, utilizing the information from the 2D images. Specifically, we render the trained SDF at each training viewpoint,  calculate the error map using the L1 distance against the inputs, sequentially binarize the map and dilate it to a candidate region $M_e$. To locate the beginning points of our growth method, we employ \cite{zhou2019end} to detect the line endpoints and dilate them to our selected region $M_s$. 
We iterate through the following process: expand $M_s$ and take the intersected set to form a new $M_s$, adding it into a mask list for training, and sequentially form a new $M_e$ by removing the updated $M_s$ region. After these preparations, the training is carried out one by one with the selected rays from the mask list.

\section{Experiments}
\subsection{Experimental settings}
\textbf{Baselines.} We conduct a comparative analysis between our proposed method and prominent approaches, including NeuS \cite{wang2021neus}, HF-NeuS \cite{wang2022hf}, and NeRF \cite{mildenhall2020nerf}, which represents the state-of-the-art pipeline without any supplementary information. We have consciously excluded MoNoSDF \cite{yu2022monosdf}, GeoNeuS \cite{fu2022geo}, and NeuralWarp \cite{darmon2022improving} from the comparison, as these methods either introduce additional priors or employ constraints from multiple viewpoints, which are similarly applicable to our method. To qualitatively evaluate the performance, we adopt two criteria: the PSNR (Peak Signal-to-Noise Ratio) for gauging novel-view rendering quality, and the Chamfer distance for accessing the accuracy of the reconstructed mesh. We obtain the underlying mesh through the marching cubes algorithm on a grid with a resolution of 1500. For NeRF, following the approach used in NeuS, we extracte the mesh using a threshold value of 25 for density.

\begin{table*}[h!]
\footnotesize
\centering
\begin{tabular}{l|ccccccccccccccc|c}
  \multicolumn{17}{c}{Chamfer Distance}  \\ 
      &24 & 37 & 40 & 55 & 63 & 65 & 69 & 83 & 97 & 105 & 106 & 110 & 114 & 118 & 122 & Mean \\ \hline
NeuS   &\cc{2}0.828 &\cc{2} 0.983 &\cc{2}0.572 & \cc{1}0.369 &\cc{2}1.185 &\cc{1}0.716 &\cc{1}0.608 & \cc{2}1.413 & 0.964 & 0.821 & \cc{1}0.495 & \cc{2}1.362 & \cc{1}0.352 & \cc{2}0.462 & \cc{1}0.499 & \cc{2}0.775 \\
NeRF   & 1.418 & 1.611 & 1.665 & 0.799 & 1.856 & 1.288 & 1.203 & 1.603 & 1.645 & 1.113 & 0.947 & 2.101 & 0.977 & 1.027 & 0.918 & 1.345 \\
HF-NeuS & 1.113 & 1.276 & 0.609 & \cc{2}0.465 & 0.973 & \cc{2}0.682 &\cc{2} 0.619 & 1.344 & \cc{1}0.914 &\cc{2} 0.728 & \cc{2}0.534 & 1.816 & \cc{2}0.378 & \cc{1}0.536 & \cc{2}0.510 &0.833 \\
Ours   & \cc{1}0.652 & \cc{1}0.913 & \cc{1}0.373 & 0.482 & \cc{1}1.049 & 0.869 & 0.821 & \cc{1}1.216 & \cc{2}0.954 & \cc{1}0.693 & 0.564 & \cc{1}1.301 & 0.416 & 0.584 & 0.569 &  \cc{1}0.764 \\

 \multicolumn{17}{c}{PSNR} \\ 
      & 24 & 37 & 40 & 55 & 63 & 65 & 69 & 83 & 97 & 105 & 106 & 110 & 114 & 118 & 122 & Mean \\ \hline
NeuS   & 27.021 & 26.602 & 27.602 & 27.651 & 35.166 & 32.119 &29.938 & 38.471 & 31.028 & 34.914 &34.638 & 33.018 & 29.888 & 37.143 & 37.764 & 32.198 \\

HF-NeuS & 28.497 & \cc{2}27.132 & \cc{2}28.986 &\cc{2} 30.554 &34.442 & \cc{1}32.892 & \cc{1}30.339 &\cc{2}38.618 & 31.014 & 35.086 &35.309 & 27.539 & 30.284 & \cc{2}37.525 & \cc{2}38.407 & 32.442 \\

NeRF   &\cc{2} 29.564 & 26.608 & 28.351 & 29.537 &\cc{2}35.838 &\cc{2}32.853 & 29.941 & 38.576 & \cc{2}31.225 & \cc{2}35.389 &  \cc{2}36.324 &  \cc{2}33.504 &  \cc{2}30.379 & 37.332 & 38.154 &  \cc{2}32.905 \\
Ours   & \cc{1}30.489 &\cc{1}27.325 &\cc{1}30.052 & \cc{1}31.387 &\cc{1}36.111 & 32.348 & \cc{2}29.985 & \cc{1}39.189 & \cc{1}31.824 & \cc{1}36.318 & \cc{1}36.519 & \cc{1}34.370 & \cc{1}31.089 &\cc{1} 38.251 & \cc{1}39.235 & \cc{1}33.633
\end{tabular}
\caption{Quantitative results on the DTU dataset. Red and orange indicate the first and second best performing results.}
\vspace{-0.2in}
\label{tab:metric_dtu}
\end{table*}

\begin{table*}
\centering
\small
\begin{tabular}{l|cccccc|c|cccccc|c}
 & \multicolumn{6}{c}{Chamfer Distance} &  & \multicolumn{6}{c}{PSNR} \\ 
      & Chair & Ficus & Lego & Materials & Mic & Ship & \textbf{Mean} & Chair & Ficus & Lego & Materials & Mic & Ship & \textbf{Mean} \\ \hline
NeuS   & 1.350 & 0.121 & 0.143 & \cc{1}0.103 & \cc{2}0.364 & 0.758 & \cc{2}0.473 & 28.590 & 25.234 & 29.348 & 29.197 & \cc{2}29.998 & 26.412 & 28.130 \\
HF-NeuS & \cc{2}0.531 & \cc{2}0.074 & \cc{2}0.070 & 0.113 & 1.971{*} & \cc{2}0.558 & 0.553 & \cc{2}28.750 & \cc{2}26.173 & 3\cc{2}0.111 & 29.448 & 29.823 & \cc{2}26.755 & \cc{2}28.514  \\
NeRF   & 1.501 & 3.660 & 1.299 & 1.069 & 1.396 & 4.272 & 2.199 & 28.196 & 25.545 & 28.474 & \cc{1}30.882 & 27.074 & 26.644 & 27.803 \\
Ours   & \cc{1}0.502 &\cc{1} 0.063 & \cc{1}0.038 & \cc{1}0.039 & \cc{1}0.094 & \cc{1}0.368 & \cc{1}0.184 & \cc{1}30.468 & \cc{1}26.713 & \cc{1}31.488 & \cc{2}30.710 & \cc{1}34.107 & \cc{1}28.360 & \cc{1}30.308
\end{tabular}
\caption{Quantitative results on the NeRF-synthetic dataset. The "Mic" case for HF-NeuS is marked with an asterisk (*) as this metric is affected by the undesired reconstruction of mesh parts, as illustrated in Fig. \ref{fig:comp_nerf}.}
\vspace{-0.2in}
\label{tab:metric_nerf}
\end{table*}

\noindent\textbf{Dataset. } 
Following the setting of previous work, we report the metric on both the DTU dataset \cite{aanaes2016large} and the NeRF-synthetic dataset. DTU is the multi-view stereo dataset. Each scene supplies 49 or 64 images with a resolution of $1600 \times 1200$, captured from various viewpoints. We adopt the foreground masks provided by IDR \cite{yariv2020multiview} for these scenes. Additionally, we conduct further testing on 7 challenging scenes from the NeRF-synthetic dataset \cite{mildenhall2020nerf}, rendering 100 images each with resolution $800 \times 800$ of black background, {without the foreground mask}. We designate every eighth image as the testing set, while the others as the training set.

\noindent\textbf{Implementation details.} In the following experiment, we select $L=5, N=6$, including planes of resolution $\{128, 256, 512, 1024, 2048\}$ and the Gaussian Kernel is of size $\{1, 1, 1, 3, 5\}$. We train our model for $300,000$ iterations, with $512$ rays randomly selected during each iteration. To ensure a fair and competitive comparison with NeuS, we maintain almost the same settings for our model.

\subsection{Comparison.}
Tab. \ref{tab:metric_dtu} and Fig. \ref{fig:comp_dtu} demonstrate quantitative and qualitative comparisons of our method against baseline methods, respectively. The results demonstrate that our method surpasses the baseline methods on the DTU dataset. Although the reported Chamfer distance does not exhibit a significant improvement, the qualitative comparison and PSNR clearly illustrate that our model can reproduce finer details. We believe this is attributed to the fact that the DTU ground truth point cloud is relatively coarse, and thus, enhancements in high-frequency details are not reflected in the metric. Additionally, the ground truth point cloud lacks some parts, which increases the distance between our reconstructed mesh and the ground truth. For further discussion, please refer to our supplementary materials. Fig. \ref{fig:comp_dtu} also showcases the incredible details that can be reproduced by our method, like the intricate details and embossed patterns on the sculptures (left), as well as the uneven surface created by the roof tiles (right), as illustrated in zoom-in details. This improvement can be demonstrated by the PSNR metric, in which our method outperforms the others, confirming the superior performance of our approach.

\begin{figure*}[tb]
    \centering
    \includegraphics[width=1\linewidth]{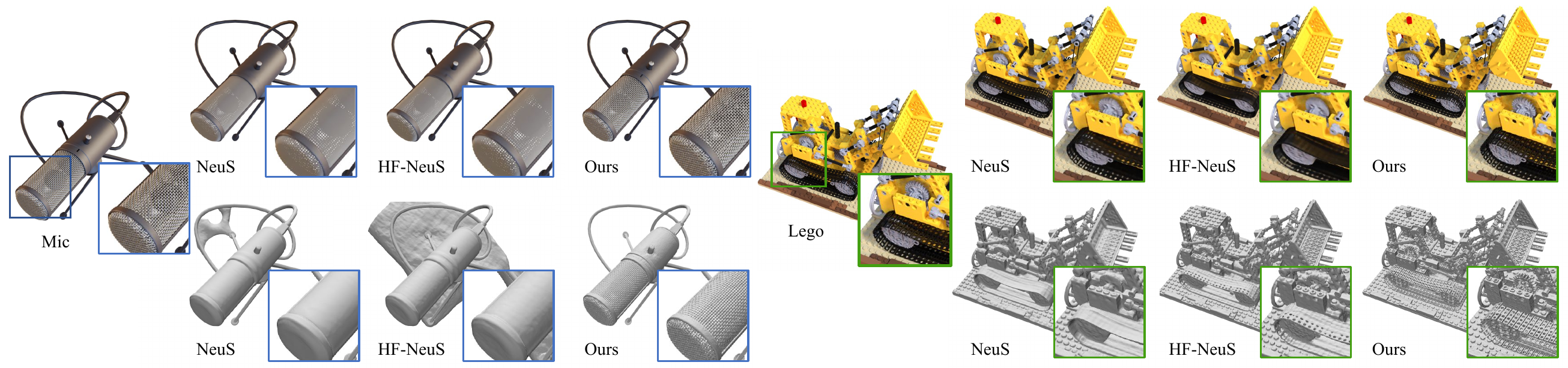}
    \caption{ Comparison of the novel-view synthesis and the reproduced mesh. We derive these detailed meshes with the marching cube of grid 1500.  Our meshes show better details on both the microphone's grille and the crawler belt of lego. Although HF-NeuS attempted to capture the high-frequency details, it struggles with geometries featuring rapid changes, such as the open hole on Lego's band. Simultaneously, the microphone's electric wire is a relatively thin object. both HF-NeuS and NeuS fail to reproduce visually pleasing meshes, as they introduce extra mesh sections connected to other parts and color them to match the background. As a result, while these methods produce appropriate novel views, the inherent geometry contains inaccuracies. Our method excels not only in reproducing details but also in maintaining smoothness on the microphone's handle, thanks to our learnable multi-level encoding method. }
    \vspace{-0.1in}
    \label{fig:comp_nerf}
\end{figure*}

\begin{figure*}[tb]
    \centering
    \includegraphics[width=0.7\linewidth]{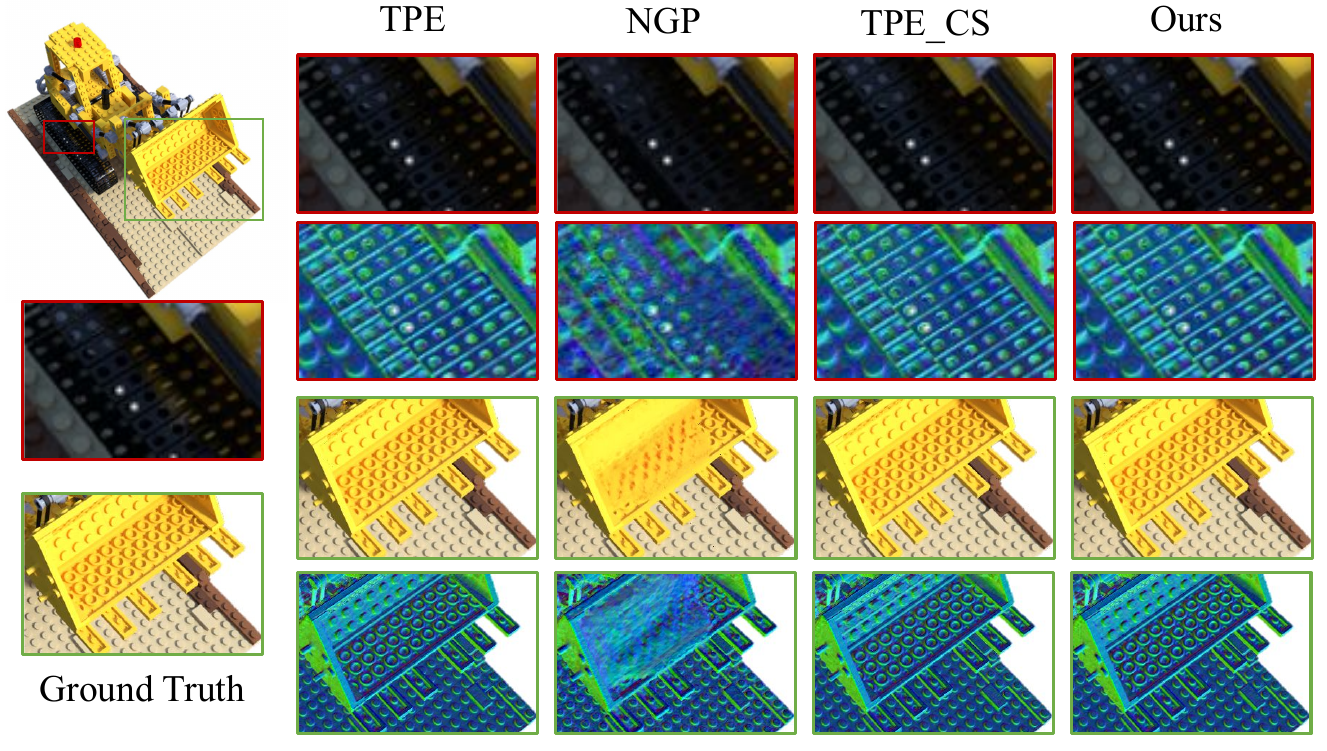}
    \caption{  We visualized the effects of different modules through novel-view synthesis, along with the corresponding surface normals. The 'NGP' encoding introduces undesired structures in regions with scarce textures or observations (e.g., the Lego’s bucket). As we progress from 'TPE' and 'TPE\_CS' to our method, we obtain increasingly detailed and clearer results, as exemplified by the distinct hole in the Lego's crawler belt.}
    \label{fig:cmp_ablation}
\end{figure*}

\begin{figure*}[tb]
    \centering
    \includegraphics[width=0.7\linewidth]{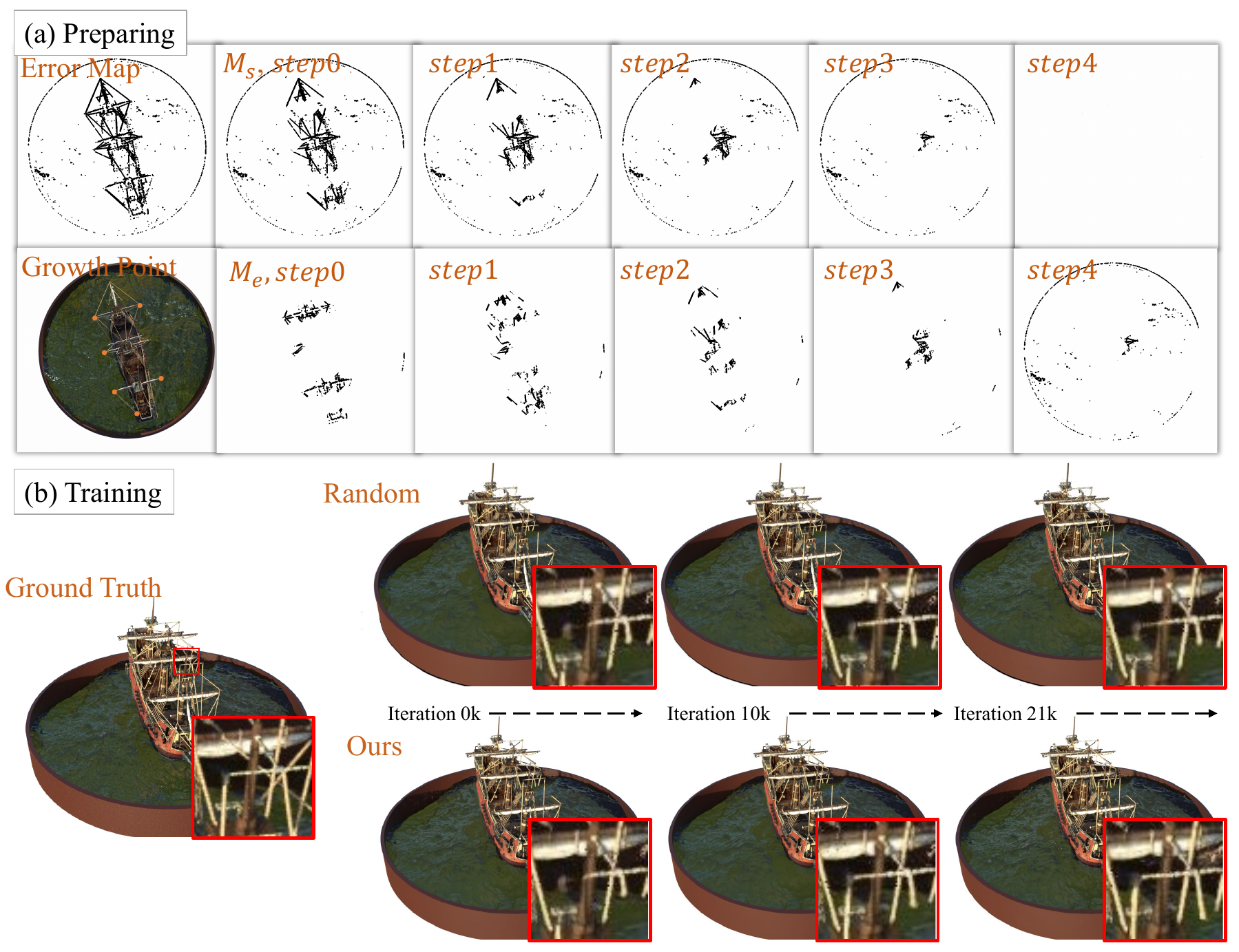}
    \caption{We compare our SDF growth method ("Ours") with randomly selected rays around the error map ("Random"). In the preparation stage, we render SDF at each training view and compute the error map ("Error Map"), starting the region growth at the detected points ("Growth Point"). We apply the region growth as Alg. \ref{alg1} with four steps for visualization. During training, the rays of $M_s$ are selected as the training set. Comparing zoom-in results, our method achieves faster convergence relative to "Random" and helps minimize the effects on regions outside the error map during refinement.
    }
    \vspace{+0.1in}
    \label{fig:refinement}
\end{figure*}

We also construct a comparison on 9 challenging models from NeRF-synthetic dataset \cite{mildenhall2020nerf}, which comprises objects exhibiting a higher level of high-frequency details. As depicted in Tab. \ref{tab:metric_nerf}, with the evaluation based on the ground truth meshes, the Chamfer distance reasonably demonstrates the superiority of our method, in contrast to the less accurate valuation on the DTU dataset. 
We provide a visual comparison in Fig. \ref{fig:comp_nerf}. 
Our method demonstrates superior performance in reproducing finer details, as evidenced by the zoom-in images of both the microphone's grille and the Lego crawler belt, compared to other approaches. HF-NeuS and NeuS struggle with geometries featuring rapid changes, partly because they employ a fixed frequency of encoding and disregard the imaging model. Moreover, as the microphone's electric wire is a relatively thin object, both HF-NeuS and NeuS fail to produce visually pleasing meshes. These methods introduce extra mesh sections connected to other parts and color them to match the background, resulting in accurate novel views but inherently inaccurate geometry. In contrast, our method not only excels in capturing rapid detail variation but also ensures smoothness on the microphone's handle, thanks to our learnable multi-level encoding approach.

\subsection{Ablation Study.}
\begin{table}
\centering
\begin{tabular}{l|ccccc}
 & NeuS  & TPE  & NGP & TPE\_CS & Ours \\ \hline
Chamfer Distance & 0.298  & 0.121  & 0.691 &\cc{2} 0.116 & \cc{1}0.114 \\
PSNR & 28.375 & 29.945  & 27.977 & \cc{2}29.951 & \cc{1}30.023 
\end{tabular}
\caption{Comparison of Chamfer Distance and PSNR in Ablation Study. }
\vspace{-0.2in}
\label{tab:metric_ablation}
\end{table}

We developed our model building upon NeuS and retained most of its settings. To evaluate our proposed modules, we performed the following comparisons:

- TPE: Replacing Plane Encoding (PE) with Multi-scale Tri-plane Encoding;

- NGP: Replacing Plane Encoding (PE) with Multi-level Grid Compression via Hashing Encoding;

- TPE\_CS: Adding Cone Sampling to the model with Multi-scale Tri-plane Encoding;

- Ours: Combining Cone Sampling with Gaussian Convolution with Multi-scale Tri-plane Encoding.

In the experiment NGP, we set $L=16, F=2, T=19, N_{min}=16$ as specified in \cite{muller2022instant} and fix $b=2$ to define the level growth factor as 2 to provide a fair basis for assessment. We conducted the comparison on the last five cases in the NeRF-synthetic dataset.The quantitative mean values and qualitative results are reported in Tab. \ref{tab:metric_ablation} and Fig. \ref{fig:cmp_ablation}, respectively.  
In the "NGP" experiment, encoding changes led to performance degradation, likely due to hash collisions from NGP's hashing mapping. This works for NeRF's low-order volume but not for SDF, where spatial points record surface distance. Thus, artifacts emerge in areas with scarce textures or observations. Fig. \ref{fig:cmp_ablation} shows this on the Lego's bucket, as observed in other research \cite{liang2023hrneus} that employed extra regularization terms to fix it.

\subsection{Efficiency of "Cone Discrete Sampling".}
Through multi-convolved featurization, our method approximates cone tracing on tri-planes. Compared to 4x super-sampling (evaluating multiple rays through each pixel and averaging them), our method only requires 1/4 of MLP queries, significantly reducing the computational burden. As shown in Tab \ref{tab:time_consuming}, our method achieves computational efficiency and superior performance. Our method takes around 160s for 1600x1200 image inference and 9h for 300k iteration training on an A100, similar to NeuS but nearly half the time of HF-NeuS. 
\begin{table}[tb]
\centering
\begin{tabular}{l|ccc}
         & TPE & super-sampling+TPE & Ours \\ \hline
GPU Memory         & 12G             & 29G      &\cc{1}13G  \\ 
Chamfer Distance   & 0.121           & 0.117    &\cc{1}0.114 \\ 

\end{tabular}
\caption{{Comparison of GPU memory, and Chamfer distance.}}
\vspace{-0.2in}
\label{tab:time_consuming}
\end{table}

\subsection{Evaluation of SDF Growth Refinement.}

We demonstrate our method that guides the SDF to converge through error-map-based guidance, as shown in Fig. \ref{fig:refinement}. We set the number of steps to $n=14$ rained for 1500 iterations at each step, resulting in a total of $21,000$ iterations of refinement for a trained model.
Our method demonstrates faster convergence than the random sampling approach. {Our SDF growth refinement enhances PSNR from $28.360$dB to $28.427$dB and reduces Chamfer distance from $0.368$ to $0.359$. This operation only adds a few seconds for initialization, making it highly efficient.} Due to computational resource limitations and dataset constraints, we only tested this module on a single, challenging case. However, we believe that this innovative method can be explored and applied to other cases, potentially leading to further improvements in a broader range of scenarios.

\section{Conclusion}
In this paper, we present a method, \textit{LoD-NeuS}, that encodes features with a continuous level of detail (LoD) from a novel tri-plane-based representation to adaptively reconstruct high-fidelity geometry. Specifically, we present a multi-scale tri-plane position encoding to capture different LoDs. To effectively represent the high-frequency sampling, we design a multi-convolved featurization to approximate the ray integral within a cone, and then aggregate LoD features from multi-convolved multi-scale features of vertices within a conical frustum along a ray. Besides, for thin surfaces, we develop an SDF growth refinement according to SDF sphere tracing for reconstruction improvement. The state-of-the-art results demonstrate the value of representing a continuous LoD to address aliasing concerns in advanced neural surface reconstruction. 

\begin{table*}[t]
    \centering
    \begin{tabular}{l|ccccccccccccccc|c}
         &24 & 37 & 40 & 55 & 63 & 65 & 69 & 83 & 97 & 105 & 106 & 110 & 114 & 118 & 122 & Mean \\ \hline
         NeuS & 1.37 & \cc{2}1.21 & 0.73 & \cc{2}0.40 & 1.20 & \cc{2}0.70 & 0.72 & \cc{2}1.01 & \cc{2}1.16 & 0.82 & 0.66 & 1.69 & 0.39 & \cc{1}0.49 & 0.51 & 0.87 \\
         NeRF & 1.90 & 1.60 & 1.85 & 0.58 & 2.28 & 1.27 & 1.47 & 1.67 & 2.05 & 1.07 & 0.88 & 2.53 & 1.06 & 1.15 & 0.96 & 1.49\\
         HF-NeuS & \cc{2}0.76 & 1.32 & \cc{2}0.70 & \cc{1}0.39 & \cc{2}1.06 & \cc{1}0.63 & \cc{2}0.63 & 1.15 & \cc{1}1.12 & \cc{2}0.80 & \cc{1}0.52 & \cc{1}1.22 & \cc{2}0.33 & \cc{1}0.49 & \cc{2}0.50 & \cc{2}0.77 \\
         Ours & \cc{1}0.69 & \cc{1}0.88 & \cc{1}0.47 & 0.42 & \cc{1}0.85 & 0.94 & \cc{1}0.59 & \cc{1}0.80 & 1.31 & \cc{1}0.64 & \cc{2}0.61 & \cc{2}1.27 & \cc{1}0.29 & \cc{2}0.64 & \cc{1}0.38 & \cc{1}0.72\\
    \end{tabular}
    \caption{Quantitative results on the DTU dataset. Red and orange indicate the first and second best-performing results.}
    \label{tab:comp}
\end{table*}

\section{Acknowledge}
This work was supported by the National Key Research and Development Program of China under Grant 2022YFF0902201, the National Natural Science Foundation of China under Grants 62001213, 62025108, and the Tencent Rhino-Bird Research Program. We thank the anonymous reviewers for their valuable feedback. 



\appendix
\section{Overview}
The supplementary material provides the implementation details  (Section \ref{sec:exp}) and the derivation of initilization of tri-plane encoding (Section \ref{sec:init}). We have also prepared a video and a reconstructed model for additional visualizations, please see the attachment.

\subsection{Experiment Setting.}
\label{sec:exp}
\textbf{Network Architecture. }
We adopt a network architecture similar to NeuS. The geometry network modeling SDF comprises 8 hidden layers with a hidden size of 256, and a skip connection concatenates the input with the output of the fourth layer. The geometry network output includes $\{1,256\}$, representing the predicted SDF and the features for the color network. Sequentially, the color network takes this feature along with the spatial position, view direction and normal to predict the point's color. As in SAL \cite{atzmon2020sal}, we employ weight normalization to the network to stabilize the training process.

\textbf{Training details.} 
We train our networks with ADAM optimizer. 
The learning rate is linearly warmed up from $0$ to $5\times10^{-4}$ in the first 5k iterations and then controlled by a cosine decay schedule to reach a minimum learning rate $2.5\times10^{-5}$.
With a ray batch size of 512, we train our network for 300k iterations, taking around 9 hours on a single Nvidia A100 GPU.
Our module only introduces slight more computation compared to NeuS. We follow the setting of Hierarchical Sampling as \cite{wang2021neus}. We evaluate our SDF growth module on the "ship" case using a learning rate of  $5\times10^{-5}$ for 21k iterations.

\subsection{Geometric Initialization of Multi-scale Tri-plane.}
\label{sec:init}
As demonstrated by the previous work \cite{atzmon2020sal}, a proper initialization is crucial to the training stabilization. However, the recent progress \cite{yu2022monosdf, liang2023hrneus} of introducing explicit voxel from NeRF to SDF has rare discussion. The crute initialize the features of the grid using the normal distribution or uniform distribution, which may make the SDF with bad initialization and failed in some textureless area.

\textbf{Initialization of Grid representation.}
We propose an initialization scheme for grid features $G$ of dimension $n$ at position $\mathbf{x}=\{x,y,z\}$. The initialized grid features consist of $\{g_i\sim\mathcal N(0,\sigma^2)\}_i^n$. In this case, the variance  $\sigma^2$ should be defined as $\sigma^2=x^2+y^2+z^2$.

\textbf{Proof}: According to SAL, an MLP $f: \mathbb{R}^d\rightarrow\mathbb{R}$ with geometric initialization, can be considered as $f(\mathbf{x})\approx \|\mathbf{x}\| - r$. That is, $f$ is approximately the signed distance function to a $d-1$ sphere of radius $r$ in $\mathbb{R}^d$. 
Assume the grid encoding as $\gamma(\mathbf{x})=\{\{g_i\}_i^n \}$, with all features concatenated as results. We can further derive that 
\begin{equation}
    f(\mathbf{\gamma(\mathbf{x})})\approx \| \{g_i\}_i^n \| - r,
\label{equ:1}
\end{equation}
where $\| \{g_i\}_i^n \|=\sqrt{\sum_i^n{g_i^2}}$. According to the law of large numbers, we get $\sum_i^n{g_i^2}=n\mathbb{E}(g)=n\sigma^2$. If we want to maintain a spatial sphere in $\mathbf{x}\in\mathbb{R}^3$ after this encoding, we should set $n\sigma^2=\|\mathbf{x}\|$, sequentially, $\sigma^2=\|\mathbf{x}\| / n$.

\textbf{Initialization of tri-plane representation.}
Extended this scheme to the case of tri-plane, which consist of three planes reflect the projection to $\{G_{xy},G_{yz},G_{zx}\}$. So we formulate $g_i=g_i^{xy}+g_i^{yz}+g_i^{zx}$, where $g_i^{xy}\sim \mathcal{N}(0,\sigma^2_{xy})$, 
$g_i^{yz}\sim \mathcal{N}(0,\sigma^2_{yz})$, 
$g_i^{zx}\sim \mathcal{N}(0,\sigma^2_{zx})$,  so $g_i \sim \mathcal{N}(0,\sigma^2_{xy}+\sigma^2_{yz}+\sigma^2_{zx})$.
Then the Equation \ref{equ:1} can be broke down as:
\begin{equation}
\begin{aligned}
   n(\sigma^2_{xy}+\sigma^2_{yz}+\sigma^2_{zx})&=\|\mathbf{x}\|=x^2+y^2+z^2
   \\&=\frac{1}{2}((x^2+y^2) + (y^2+z^2) + (x^2+z^2)).
\end{aligned}
\end{equation}
Therefore, we initialize the three plane as $\sigma^2_{xy}=\frac{1}{2n}(x^2+y^2),\sigma^2_{yz}=\frac{1}{2n}(y^2+z^2),\sigma^2_{zx}=\frac{1}{2n}(z^2+x^2)$.

\subsection{Comparison}
We compared our method with the state-of-the-art methods (\textit{e.g.} NeRF \cite{mildenhall2020nerf}, NeuS \cite{wang2021neus}, HF-NeuS \cite{wang2022hf}) on the DTU dataset without mask supervision, as shown in Tab. \ref{tab:comp}. Our method achieves an average Chamfer distance of 0.72 v.s. 0.77 (from HF-NeuS's paper), demonstrating our method still exhibits superior performance. Both metrics improve compared to those with masks because the peripheries of meshes are masked before evaluation, following the post-processing in HF-NeuS.

\bibliographystyle{ACM-Reference-Format}
\bibliography{sample-bibliography}

\end{document}